\documentclass{article}

\usepackage[final]{AdvML_Frontiers_2024}

\usepackage[utf8]{inputenc} 
\usepackage[T1]{fontenc}    
\usepackage{url}            
\usepackage{booktabs}       
\usepackage{amsfonts}       
\usepackage{nicefrac}       
\usepackage{microtype}      
\usepackage{xcolor}         

\usepackage{graphicx}
\usepackage{algorithm}
\usepackage{algorithmic}
\usepackage{multirow}
\usepackage{booktabs}
\usepackage{amsmath} 
\usepackage{enumitem}

\definecolor{cvprblue}{rgb}{0.21,0.49,0.74}
\definecolor{cvprred}{rgb}{0.74,0.21,0.49}
\usepackage[breaklinks,colorlinks,citecolor=cvprblue,linkcolor=cvprred]{hyperref}

\title{Adversarial Bounding Boxes Generation (ABBG) Attack against Visual Object Trackers}

\author{%
  Fatemeh Nourilenjan Nokabadi \\ 
  Université Laval, IID, Mila\\
  \texttt{fatemeh.nourilenjan-nokabadi.1@ulaval.ca} \\
  \AND
  Jean-Fran\c{c}ois Lalonde\\
  Université Laval, IID\\
  \texttt{jean-francois.lalonde@gel.ulaval.ca}\\
  \And
  Christian Gagné \\
  U. Laval, IID, Mila, Canada-CIFAR AI Chair\\
  \texttt{christian.gagne@gel.ulaval.ca} \\
}

\begin{document}

\maketitle

\begin{abstract}
Adversarial perturbations aim to deceive neural networks into predicting inaccurate results. For visual object trackers, adversarial attacks have been developed to generate perturbations by manipulating the outputs. However, transformer trackers predict a specific bounding box instead of an object candidate list, which limits the applicability of many existing attack scenarios. To address this issue, we present a novel white-box approach to attack visual object trackers with transformer backbones using only one bounding box. From the tracker predicted bounding box, we generate a list of adversarial bounding boxes and compute the adversarial loss for those bounding boxes. Experimental results demonstrate that our simple yet effective attack outperforms existing attacks against several robust transformer trackers, including TransT-M, ROMTrack, and MixFormer, on popular benchmark tracking datasets such as GOT-10k, UAV123, and VOT2022STS.
\end{abstract}

\section{Introduction}

Visual object trackers with transformer backbones have consistently ranked among the most robust models in tracking benchmarks~\citep{kristan_tenth_2023}. By utilizing cross-attention and self-attention layers, transformer trackers~\citep{cui_mixformer_2022, chen_high-performance_2023, cai_robust_2023} infer a single bounding box per frame by jointly extracting features from both the search and template regions. However, the adversarial robustness of transformer trackers has yet to be thoroughly studied~\citep{nokabadi2024reproducibility}. White-box attacks~\citep{guo_spark_2020, jia_robust_2020} against object trackers focus on manipulating the tracker’s output in a way that the adversarial loss generates perturbed frame patches by backpropagating through the network. The applicability of such attacks depends on the attack proxy, i.e., the element being perturbed by the attack mechanism, within the white-box setting. For example, RTAA~\citep{jia_robust_2020} generates perturbed search regions using the classification and regression labels inferred by the object trackers. It is therefore not applicable on trackers which do not rely on such classification and regression labels.The efficiency of white-box attacks is generally greater than that of black-box attacks~\citep{nokabadi2024reproducibility}, as white-box methods leverage the tracker’s gradients to generate perturbations. However, there is currently no white-box attack applicable to a wide range of new transformer trackers that allows for a comprehensive comparison of their adversarial robustness. Some attacks rely on classification and regression labels (such as SPARK~\citep{guo_spark_2020} and RTAA~\citep{jia_robust_2020}), while others require heatmaps, CSA~\citep{yan_cooling-shrinking_2020}, to be effective as white-box attacks. These attack proxies are available in some tracking frameworks, such as TransT~\citep{chen_transformer_2021}, but are not available in other tracking pipelines, such as ROMTrack~\citep{cai_robust_2023}. To address this gap, we propose the Adversarial Bounding Box Generation (ABBG) attack, which uses only a single bounding box to challenge the object tracker’s robustness against adversarial perturbations in a white-box setting. Our method is simple yet efficient and is applicable to a wide range of transformer trackers.

\begin{figure}
    \centering     
    \centerline{\includegraphics[width=0.8\textwidth]{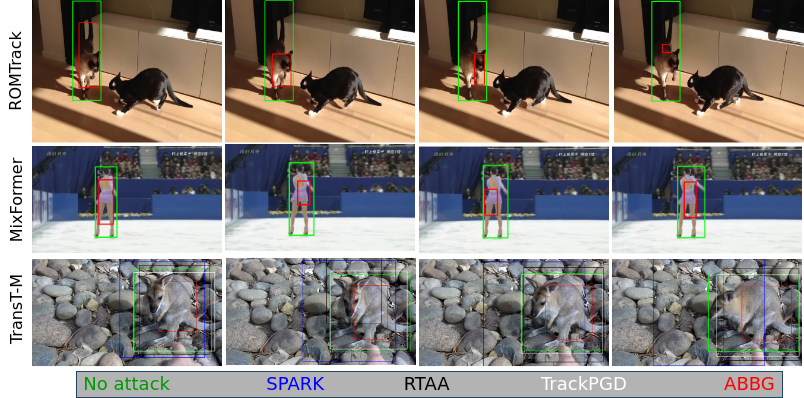}}
 \caption{The adversarial robustness of transformer-based trackers, including ROMTrack~\citep{cai_robust_2023}, MixFormer~\citep{cui_mixformer_2022}, and TransT-M~\citep{chen_high-performance_2023}, is evaluated against white-box attacks, such as SPARK~\citep{guo_spark_2020} (blue), RTAA~\citep{jia_robust_2020} (black), TrackPGD~\citep{nokabadi24TPGD} (white), and our proposed ABBG attack (red). Our proposed ABBG method generates adversarial perturbations by manipulating only the target prediction of the bounding box. The ABBG attack is applicable as a white-box attack to a wide range of transformer-based trackers, including MixFormer and ROMTrack, while other white-box attacks, such as RTAA, SPARK, and TrackPGD, are not applicable due to the unavailability of an attack proxy.}
     \label{fig:cont}
\end{figure}

Figure~\ref{fig:cont} demonstrates the performance of ABBG in misleading transformer trackers, e.g. ROMTrack~\citep{cai_robust_2023}, MixFormer~\citep{cui_mixformer_2022} and TransT-M~\citep{chen_high-performance_2023} in predicting target bounding boxes after applying white-box attacks including SPARK~\citep{guo_spark_2020}, RTAA~\citep{jia_robust_2020}, TrackPGD~\citep{nokabadi24TPGD}, and ABBG (our method). Our method uses the single predicted bounding box of the trackers as the attack proxy. Because of this, it is the first white-box attack that is applicable to a wide range of trackers with transformer backbones. 
In summary, the contributions of our work are:
\begin{itemize}[nosep]
    \item We propose a white-box attack, applicable on a wide range of transformer trackers, which manipulates a single bounding box predicted by the tracker.
    \item In affecting transformer trackers, ABBG attack is ranked first in several tracking datasets per at least one evaluation metric, while in terms of sparsity and imperceptibility of generated perturbations, our method is ranked second in comparison to other competitive white-box attacks.
\end{itemize}

\section{Related works}



The visual object tracking (VOT) task has been studied extensively in the computer vision community, for instance in real-time scenarios, data-driven models, online/offline tracking, etc. The use of vision transformers in the backbone of tackers let to a brand new category of object trackers which are listed as the most robust trackers in many tracking assessments and datasets~\citep{huang_got-10k_2021, kristan_tenth_2023}. In transformer trackers~\citep{cui_mixformer_2022,chen_high-performance_2023,chen_transformer_2021,cai_robust_2023}, the main difference concerns how cross-attention and self-attention layers are defined and developed. In TransT~\citep{chen_transformer_2021}, a multi-head pipeline similar to traditional Siamese-based trackers is adopted while two modules of self-attention and cross attention named Ego-Context Augment (ECA) and Cross-Feature Augment (CFA), respectively, are proposed to process search and template regions. Built upon TransT~\citep{chen_transformer_2021}, the TransT-SEG and TransT-M~\citep{chen_high-performance_2023} are introduced with segmentation ability. The OSTrack~\citep{ye_joint_2022} proposed a one-stream joint feature extraction and relation modeling for template and search regions. The in-network early candidate elimination module of OSTrack improves the tracking performance by gradually identifying and discarding the background regions. The MixFormer tracker~\citep{cui_mixformer_2022} introduces the Mixed Attention Module (MAM) to deeply integrate the self- and cross-attention layers. A score prediction module is also proposed to fuse the object candidate scores and infer the target score. In the Robust Object Modeling Tracker (ROMTrack)~\citep{cai_robust_2023} inherent and hybrid template modeling is suggested in one-stream robust modeling. Using variation tokens, the appearance changes of the target is depicted in the tracking process.


Adversarial robustness in object trackers has been explored through the development of black-box attacks, such as the IoU method~\citep{jia_iou_2021}, and white-box attacks, such as RTAA \citep{jia_robust_2020}, SPARK~\citep{guo_spark_2020}, and TrackPGD~\citep{nokabadi24TPGD}. Unlike black-box methods, white-box attacks generate adversarial perturbations by directly manipulating tracker outputs. However, due to changes in the backbone architecture of transformer trackers, several previous white-box methods, including SPARK, RTAA, TrackPGD, and CSA~\citep{yan_cooling-shrinking_2020}, are no longer applicable to these newer models in a white-box setting. The primary objective of our research is to evaluate the effectiveness of white-box attacks on transformer trackers, utilizing transformer gradients to deceive the object trackers.

\section{Proposed Method}


\begin{figure*}[t]
     \centering
     \centerline{\includegraphics[width=0.8\textwidth]{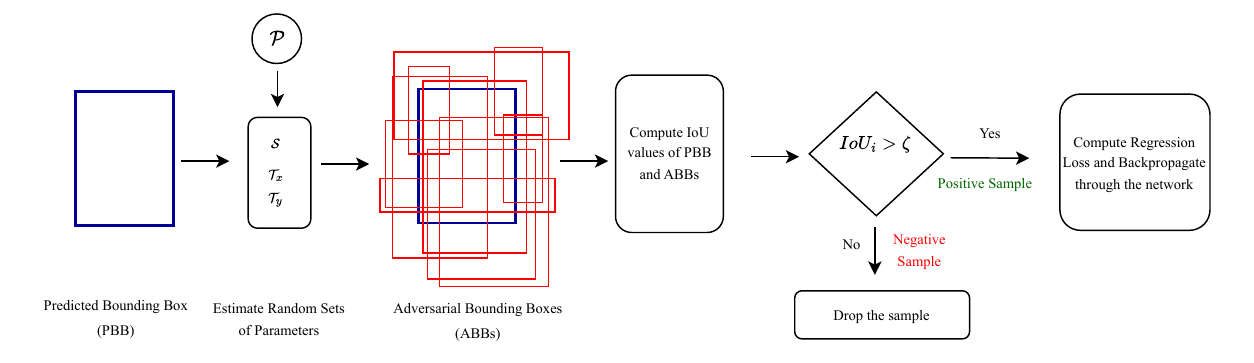}}
     \caption{The overview of the ABBG attack approach. The random sets of scale $\mathcal{S}$,'x-axis' translation $\mathcal{T}_x$ and 'y-axis' translation $\mathcal{T}_y$ are sampled from uniform distributions to apply on the predicted bounding boxes using Equation~\ref{eq:genB}. The adaptive threshold $\zeta$ is computed based on the set of obtained IoUs per each iteration step of the attack.}
     \label{fig:method}
\end{figure*}


Our goal is to mislead transformer trackers into predicting inaccurate bounding boxes across video frames. The Adversarial Bounding Boxes Generation (ABBG) attack manipulates the tracker's predicted bounding box to create adversarial search regions. Figure~\ref{fig:method} provides an overview of the ABBG attack in computing the adversarial loss, which is then backpropagated through the network to generate the adversarial search regions for transformer trackers.


Given the predicted object bounding box $[x, y, w, h]$, our attack method generates $k$ adversarial bounding boxes using random parameters as follows:
\begin{align} \label{eq:genB}
   & x'_i = x + \mathcal{T}_{x_i} \,, y'_i = y + \mathcal{T}_{y_i} \nonumber \\
   & w'_i = w s_i \,, h'_i = h s_i \,,
\end{align}
where $i \in \{ 1, 2, ..., k \}$ corresponds to each adversarial bounding box. The offset $\mathcal{T}$ and scale $s$ parameters are selected randomly from uniform distributions in each attack step. From these parameters, the $k$ number of bounding boxes are generated around the predicted bounding box, see Figure~\ref{fig:method}. The bounding boxes which are closest to the the predicted bounding box are chosen as positive samples. The selection process is based on an adaptive threshold that only retains bounding boxes with an Intersection over Union (IoU) greater than a fixed threshold, specifically $80\%$ of all adversarial bounding boxes. Adaptive thresholding is crucial because it ensures that some bounding boxes are always selected to compute the adversarial loss. The adversarial loss of the ABBG attack is is a regression loss computed between adversarial bounding boxes $b^*_i$ and the true predicted bounding box $b_{pred}$:
\begin{equation}
    \ell_\mathrm{ABBG} = \Sigma_i \ell_r (b_{pred}, b^*_{i}) \,,
\end{equation}
where $\ell_r$ the smoothed L1 norm. In our method, we fixed the number of iterations to $10$, similar to other white-box attacks, RTAA~\citep{jia_robust_2020} and SPARK~\citep{guo_spark_2020}. Also, we clip the generated perturbation in the $\epsilon$-ball where $\epsilon = 10$, as in SPARK and RTAA. 


\section{Experimental results}
\label{sec:exp}


\paragraph{Object trackers.} We assessed our proposed attack method against three transformer trackers: TransT-M~\citep{chen_transformer_2021}, MixFormer~\citep{cui_mixformer_2022}, and ROMTrack~\citep{cai_robust_2023}.

\paragraph{Datasets $\&$ metrics.} We measured the performance of attacks on three tracking benchmarks: GOT-10k~\citep{huang_got-10k_2021}, UAV123~\citep{mueller_benchmark_2016}, and VOT2022-ST~\citep{kristan_tenth_2023}. The evaluated metrics include (Expected) Average Overlap (EAO or AO), success rates at the $0.5$ and $0.75$ overlap thresholds, denoted as $SR_{0.5}$ and $SR_{0.75}$, respectively, along with accuracy and robustness under the anchor-based short-term tracking protocol~\citep{kristan_tenth_2023}, L1-norm, and structural similarity index (SSIM)~\citep{wang_image_2004}. The term "drop $\%$" refers to the percentage reduction observed in each metric.

\paragraph{Adversarial attacks.} Three white-box attacks including SPARK~\citep{guo_spark_2020}, RTAA~\citep{jia_robust_2020}, TrackPGD~\citep{nokabadi24TPGD} and two black-box attacks containing IoU~\citep{jia_iou_2021} and CSA~\citep{yan_cooling-shrinking_2020} are compared to our proposed attack results. Note that CSA~\citep{yan_cooling-shrinking_2020} is used as a transferred attack using the SiamRPN++ tracker~\citep{li_siamrpn_2019}. 

\paragraph{Attack setup.} In our experiments, the translation parameters ($\mathcal{T}{x_i}$, $\mathcal{T}{y_i}$) were sampled from uniform distributions between $0.1$ and $0.4$, with $k = 1024$. The scale parameters ($\mathcal{S}$) were sampled from a uniform distribution between $0.7$ and $0.9$, also with $k = 1024$. Following the approach of SPARK~\citep{guo_spark_2020} and RTAA~\citep{jia_robust_2020}, we set the attack's training period to $10$ steps.

\subsection{Quantitative results}

Table~\ref{tab:gtest} shows the performance of object trackers when attacked by different methods compared to our proposed ABBG attack on the GOT10k dataset~\citep{huang_got-10k_2021}. Among white-box attacks, including RTAA, SPARK, TrackPGD, and ABBG (ours), our method is applicable to all three trackers- TransT-M, ROMTrack, and MixFormer. Beyond this versatility, the ABBG attack causes significant drops in all three measured scores, including average overlap and success rates at the $0.5$ and $0.75$ thresholds on the GOT10k dataset. For MixFormer~\citep{cui_mixformer_2022} and ROMTrack~\citep{cai_robust_2023}, our ABBG attack is the first white-box attack to seriously challenge these transformer trackers by reducing their scores to nearly zero.

We also evaluate transformer tracker performance against adversarial attacks on the VOT2022-ST dataset and protocol~\citep{kristan_tenth_2023}. Table~\ref{tab:votT} presents the results of attacks using three metrics: EAO, accuracy, and robustness, based on the baseline experiment of the VOT short-term sub-challenge. The anchor-based short-term tracking protocol of the VOT2022-ST evaluation~\citep{kristan_tenth_2023} requires trackers to reinitialize whenever the target is lost for a significant number of time steps. Under this protocol, our attack performs well compared to other white-box attacks- RTAA~\citep{jia_robust_2020}, SPARK~\citep{guo_spark_2020}, and TrackPGD~\citep{nokabadi24TPGD}- on the TransT-M tracker~\citep{chen_high-performance_2023}, except in the accuracy metric, where we ranked second, following the SPARK method. Additionally, we assessed the white-box attacks on TransT-M using another dataset for further investigation. We selected TransT-M because it is the only tracker among our set where all four white-box attacks can be applied. On the UAV123 dataset~\citep{mueller_benchmark_2016}, the measured metrics, shown in Table~\ref{tab:uav}, indicate the superiority of the ABBG attack in success rate, while SPARK caused slightly more damage—around $1.23\%$—in the precision metric.

\begin{table*}
    \centering \footnotesize
    \caption{The performance of object trackers~\citep{chen_high-performance_2023, cui_mixformer_2022, cai_robust_2023} after adversarial attacks including SPARK~\citep{guo_spark_2020}, RTAA~\citep{jia_robust_2020}, TrackPGD~\citep{nokabadi24TPGD}, IoU~\citep{jia_iou_2021} and CSA~\citep{yan_cooling-shrinking_2020} and ABBG (ours), on the GOT10k~\citep{huang_got-10k_2021} dataset. The \textit{Italic} methods are white-box attacks.} 
    \label{tab:gtest}
    \begin{tabular}{p{0.1\textwidth}p{0.14\textwidth}p{0.09\textwidth}p{0.09\textwidth}p{0.09\textwidth}p{0.09\textwidth}p{0.09\textwidth}p{0.09\textwidth}} \\ 
    \toprule
     Tracker & Attacker & \multicolumn{3}{c}{Scores} & \multicolumn{3}{c}{Drop(\%)}    \\
     & & AO & SR$_{0.5}$ & SR$_{0.75}$ &  AO & SR$_{0.5}$ & SR$_{0.75}$ \\
    \midrule 
     \multirow{6}{*}{TransT-M}
      & No Attack  & 0.734 & 0.840 & 0.693&  - & - &  - \\
      & CSA   & 0.683 & 0.779 & 0.635 & 6.95 & 7.26 & 8.37 \\
      & IoU   & 0.587 & 0.672 & 0.480 & 20.03 & 20.00 & 30.74 \\
      & \textit{SPARK} & 0.116 & 0.065 & 0.023 & 84.20 & 92.26 & 96.68 \\
      & \textit{RTAA} & 0.173 & 0.180 & 0.112 & 76.43 & 78.57 & 83.84 \\
      & \textit{TrackPGD} & 0.388 & 0.429 & 0.188 & 47.14 & 48.93 & 72.87 \\
      & \textit{ABBG (ours)} & {\bf 0.027} & \bf 0.019 & \bf 0.001 & \bf 96.32 & \bf 97.74 & \bf 99.85 \\
      \midrule
      \multirow{3}{*}{ROMTrack} 
        & No Attack & 0.729 & 0.830 & 0.702 & - & - & - \\
        & CSA & 0.716 & 0.814 & 0.682 & 1.78 & 1.93 & 2.85 \\
        & IoU  & 0.597 & 0.678 & 0.536 & 18.11 & 18.31 & 23.65 \\
        & \textit{ABBG (ours)} & \bf 0.006 & \bf 0.001 & \bf 0.000 & \bf 99.18 & \bf 99.88 & \bf 100 \\
      \midrule
      \multirow{3}{*}{MixFormer}
       & No Attack & 0.696 & 0.791 & 0.656 & - & - & - \\
       & CSA   &  0.638 & 0.727 & 0.572 & 8.33 & 8.10 & 12.80 \\
      & {IoU} & 0.625 & 0.713 & 0.543 & {10.20} & {9.86} & 17.22 \\
       & \textit{ABBG (ours)} & \bf 0.002 & \bf 0.000 & \bf  0.000 & \bf 99.71 & \bf 100 & \bf 100 \\ 
       \bottomrule
    \end{tabular}
\end{table*}

\begin{table*}
    \centering  \footnotesize
    \caption{Evaluation results of the TransT-M~\citep{chen_high-performance_2023} attacked by different methods including SPARK~\citep{guo_spark_2020}, RTAA~\citep{jia_robust_2020}, TrackPGD~\citep{nokabadi24TPGD}, IoU~\citep{jia_iou_2021} and CSA~\citep{yan_cooling-shrinking_2020} and ABBG (ours), on the VOT2022-STS~\citep{kristan_tenth_2023} dataset. The reported scores are from the Baseline experiment of VOT2022-STS evaluation.The \textit{Italic} methods are white-box attacks.}
    \label{tab:votT}
    \begin{tabular}{p{0.12\textwidth}p{0.14\textwidth}p{0.08\textwidth}p{0.08\textwidth}p{0.1\textwidth}p{0.08\textwidth}p{0.08\textwidth}p{0.1\textwidth}} \\ 
    \toprule
    Tracker & Attacker & \multicolumn{3}{c}{Scores} & \multicolumn{3}{c}{Drop($\%$)}   \\
     & & EAO & Accuracy &  Robustness & EAO & Accuracy & Robustness \\
    \midrule 
    \multirow{7}{*}{TransT-M} 
       & No attack & 0.542 & 0.744 & 0.866 & - & - & -  \\
       & CSA & 0.480 & 0.719 & 0.815 & 11.44 & 3.36 & 5.89 \\
       & IoU & 0.332 & 0.654 &  0.646 & 38.74 & 12.10 & 25.40  \\
       & \textit{RTAA} & 0.052 & 0.601 & 0.088 & 90.40 & 19.22 & 89.84  \\
       & \textit{SPARK} & 0.009 & 0.261 & 0.041 & 98.34 & {\bf 64.92} & 95.26  \\
       & \textit{TrackPGD}  & 	0.172 & 0.431 & 0.491 & 68.26 & 42.07 & 43.30 \\
       & \textit{ABBG (ours)} & 0.007 & 0.370 & 0.018 & {\bf 98.71} & 50.27 & {\bf 97.92}
       \\ 
       \bottomrule
    \end{tabular}
\end{table*}

\begin{table*}
    \centering  \footnotesize
    \caption{Evaluation of our proposed ABBG attack compared to different white-box attacks including SPARK~\citep{guo_spark_2020}, RTAA~\citep{jia_robust_2020}, TrackPGD~\citep{nokabadi24TPGD} and ABBG (ours), applied on the TransT-M~\citep{chen_high-performance_2023} tracker on the UAV123~\citep{mueller_benchmark_2016} dataset.}
    \label{tab:uav}
    \begin{tabular}{p{0.12\textwidth}p{0.15\textwidth}p{0.12\textwidth}p{0.12\textwidth}p{0.12\textwidth}p{0.12\textwidth}} \\ 
    \toprule
     & & \multicolumn{2}{c}{Attack} & \multicolumn{2}{c}{Drop($\%$)}  \\
    Tracker & Attacker & Success & Precision & Success & Precision  \\
    \midrule 
    \multirow{5}{*}{TransT-M}
        & No attack & 0.697 & 0.893 & - & - \\
        & SPARK &  0.056 & 0.068 & 91.96 & {\bf 92.38}  \\
        & RTAA &  0.083 & 0.120 & 57.79 & 86.56  \\
        & TrackPGD &  0.264 & 0.441 & 26.12 & 50.61   \\
        & ABBG (ours) &  0.024 & 0.079 & {\bf 96.55} & 91.15   \\ 
       \bottomrule
    \end{tabular}
\end{table*}

\subsection{Qualitative results}

\begin{figure}
    \centering 
    \centerline{\includegraphics[width=0.8\textwidth]{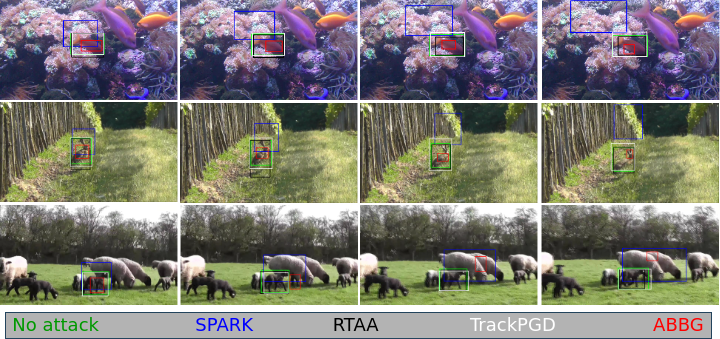}}
 \caption{Several examples of TransT-M~\citep{chen_high-performance_2023} performance after applying the white-box attacks containing SPARK~\citep{guo_spark_2020} (blue), RTAA~\citep{jia_robust_2020} (black), TrackPGD~\citep{nokabadi24TPGD} (white), and our proposed ABBG attack (red) bounding boxes. The Green color represents the tracker's original response with no attack applied.} 
     \label{fig:qr}
\end{figure}

In Figure~\ref{fig:qr} we compared the effectiveness of white-box attacks in degrading object bounding box in the TransT-M~\citep{chen_high-performance_2023} tracker on several video sequences starting from the second frame. More examples are provided in the Supplementary materials.

\subsection{Sparsity and imperceptibility}


We assess sparsity by calculating the L1 norm between the perturbed and clean search regions, while imperceptibility is evaluated using the Structural Similarity (SSIM) index~\citep{wang_image_2004} between these regions. Our experiment involves four attack methods -SPARK~\citep{guo_spark_2020}, RTAA~\citep{jia_robust_2020}, TrackPGD~\citep{nokabadi24TPGD}, and ABBG— tested against the TransT-M tracker~\citep{chen_high-performance_2023}. All attacks are conducted in a white-box setting, utilizing 10 iteration steps to generate adversarial perturbations. In each method, pixel values are allowed to vary by up to 10 units per update, ensuring a consistent budget for image manipulation. Table~\ref{tab:pertMT} indicates the sparsity and imperceptibility of adversarial perturbations generated by different white-box attacks on the VOT2022-ST dataset~\citep{kristan_tenth_2023}. Our method is ranked second in both metrics after SPARK~\citep{guo_spark_2020}; while its generated perturbations are more effective than SPARK on GOT-10k~\citep{huang_got-10k_2021} and VOT2022ST~\citep{kristan_tenth_2023}, Tables~\ref{tab:gtest} and~\ref{tab:votT}. Also, our method is applicable to ROMTrack~\citep{cai_robust_2023} and MixFormer~\citep{cui_mixformer_2022} where SPARK is not applicable as a white-box attack. 

\begin{table}
    \centering \footnotesize
    \caption{Perturbation metrics generated by various attack methods on TransT-M tracker~\citep{chen_high-performance_2023} using the VOT2022 dataset (Short-term subchallenge)~\citep{kristan_tenth_2023}.}
    \label{tab:pertMT}
    \begin{tabular}{p{0.15\textwidth}p{0.15\textwidth}p{0.15\textwidth}p{0.15\textwidth}}\\ \toprule
     Tracker & Attacker & L1-Norm $\downarrow$ & SSIM($\%$) $\uparrow$ \\ \toprule
     \multirow{4}{*}{TransT-M}
    & SPARK & 69.98 & 94.43 \\
    & RTAA & 113.48 & 60.14  \\
    & TrackPGD & 122.52 & 64.04 \\
    & {ABBG (ours)} & 95.77 & 89.50 \\ 
    \bottomrule
    \end{tabular}
\end{table}


\section{Conclusion}

In this paper, we evaluated the adversarial robustness of three transformer trackers under a novel white-box attack that manipulates target bounding box predictions to generate adversarial perturbations. Our attack, ABBG, is broadly applicable across trackers, leveraging the simplicity of a single bounding box as its attack proxy. We demonstrated that ABBG consistently outperforms other attack methods, excelling in at least one metric per dataset. Notably, ABBG ranks second in both sparsity and imperceptibility among white-box attacks after a fixed number of iterations. We are confident that this work will drive further research into adversarial vulnerabilities in transformer trackers and catalyze efforts to enhance their robustness.


\subsubsection*{Acknowledgments}

This work is supported by the DEEL Project CRDPJ 537462-18 funded by the Natural Sciences and Engineering Research Council of Canada (NSERC) and the Consortium for Research and Innovation in Aerospace in Québec (CRIAQ), together with its industrial partners Thales Canada inc, Bell Textron Canada Limited, CAE inc and Bombardier inc. \footnote{\url{https://deel.quebec}}

\small

\bibliography{bibliography}
\clearpage


\appendix

\section{Supplementary materials}

\subsection{Metric explanation}

In this subsection, we describe the calculation of several metrics related to the VOT2022 protocol~\citep{kristan_tenth_2023} to provide a clearer understanding of the results.

\paragraph{Robustness} In the VOT2022 protocol, robustness is measured using anchor-based evaluation. In this method, multiple anchors are placed in each test sequence, approximately 50 frames apart, to facilitate anchor-based short-term tracking~\citep{kristan_tenth_2023}. The tracker is initialized in the first frame, and whenever a "tracking failure" occurs, it is re-initialized from the previous or next anchor, depending on which direction has the longer remaining sequence. Robustness is defined as the length of the video subsequence during which the tracker avoids tracking failure. A tracking failure occurs when the overlap between the prediction and the ground truth falls below a certain threshold for a specified number of time steps, as defined in the VOT protocol.

\paragraph{Accuracy} The accuracy metric, as measured in the VOT2022 baseline experiment, calculates the average overlap between the prediction and the ground truth under the anchor-based short-term tracking protocol~\citep{kristan_tenth_2023}.

\paragraph{Expected Average Overlap (EAO)} The EAO metric in the VOT2022 baseline experiment follows the anchor-based short-term tracking protocol~\citep{kristan_tenth_2023}. EAO combines accuracy and robustness metrics by computing the average overlap values for subsequences up to the point of tracking failure and then reporting the overall average across all subsequences.

\subsection{Bounding box and binary mask evaluation}
\begin{figure}[h]
    \centering 
    \centerline{\includegraphics[width=\textwidth]{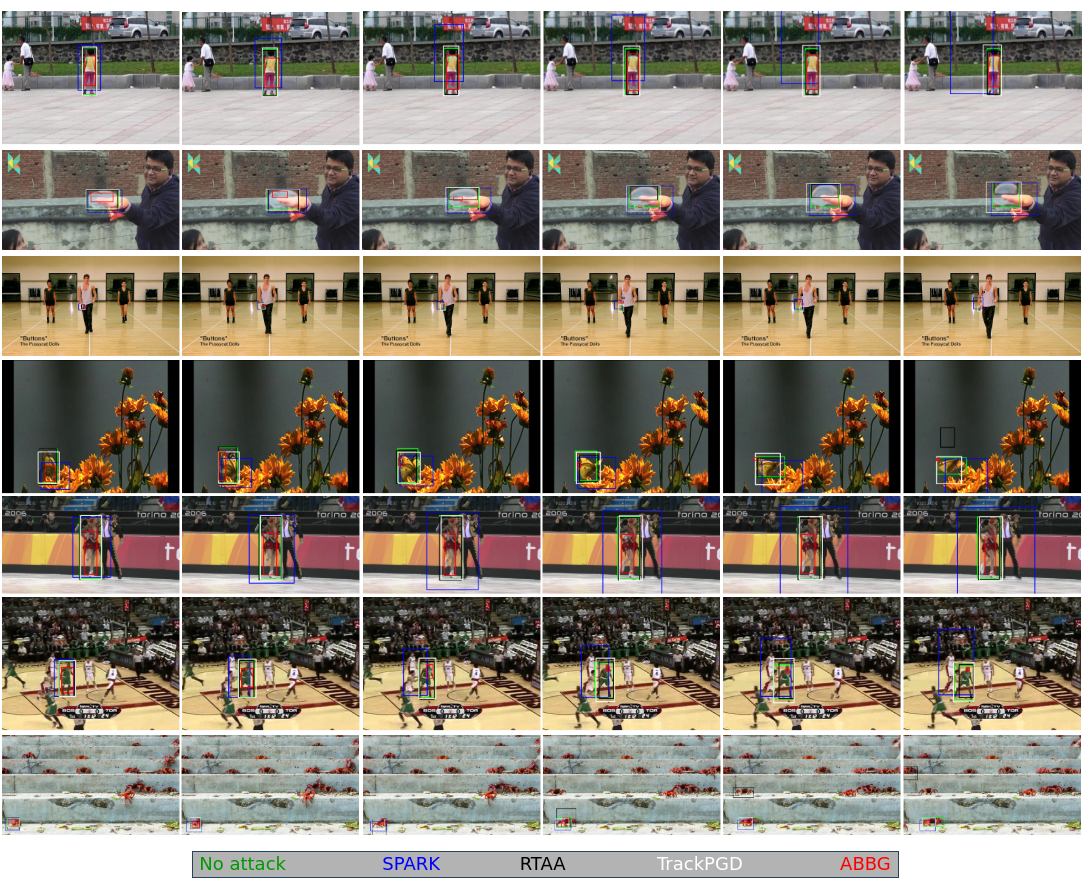}}
 \caption{Several examples of TransT-M~\citep{chen_high-performance_2023} performance after applying the white-box attacks containing SPARK~\citep{guo_spark_2020} (blue), RTAA~\citep{jia_robust_2020} (black), TrackPGD~\citep{nokabadi24TPGD} (white), and our proposed ABBG attack (red) bounding boxes. The Green color represents the tracker's original response with no attack applied.} 
     \label{fig:spb}
\end{figure}

We demonstrate the performance of our ABBG attack on video sequences, showcasing its ability to mislead object trackers for bounding box predictions, in comparison to other attack methods, as shown in Figure~\ref{fig:spb}. Several examples are included, featuring various moving objects that differ in size, color, background, and scene context.


\begin{figure}[t]
    \centering 
    \centerline{\includegraphics[width=\textwidth]{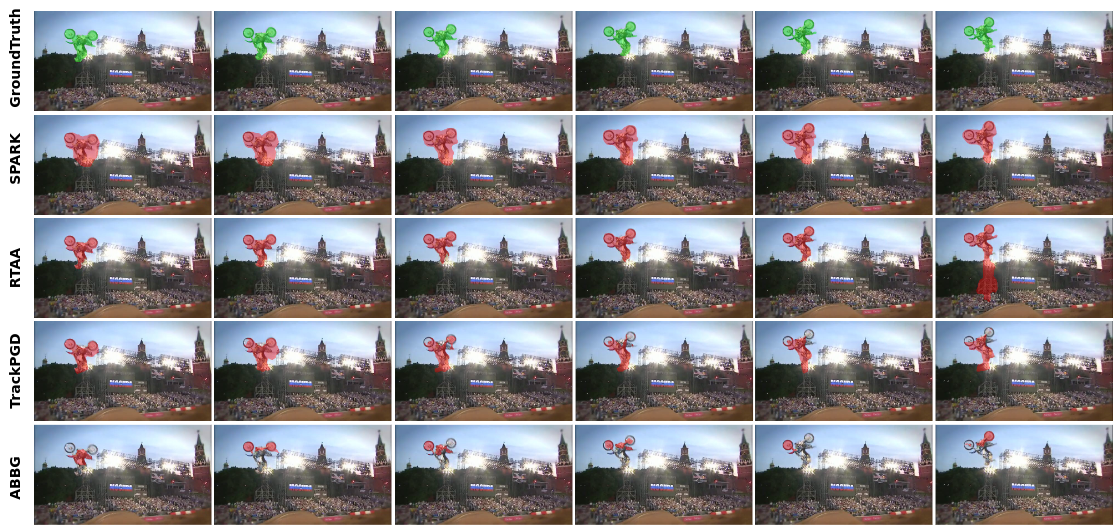}}
 \caption{Several examples of TransT-M~\citep{chen_high-performance_2023} performance after applying the white-box attacks containing SPARK~\citep{guo_spark_2020} (blue), RTAA~\citep{jia_robust_2020} (black), TrackPGD~\citep{nokabadi24TPGD} (white), and our proposed ABBG attack (red) binary masks. The Green color represents the ground truth (mask).} 
     \label{fig:spb2}
\end{figure}

The object binary mask prediction is also illustrated after applying different attack methods in Figure~\ref{fig:spb2}. Note that the "Green" mask is the original tracker response while the "Red" mask is the tracker response after applying the attack. The SPARK, RTAA and ABBG attack is the bounding box-based attack where the TrackPGD is using object binary mask in generating the adversarial perturbations. 

\end{document}